\documentclass[11pt,a4paper]{article}
\usepackage[hyperref]{conll-2019}
\usepackage{times}
\usepackage{latexsym}

\usepackage{url}

\aclfinalcopy 

\usepackage{enumitem}
\usepackage{amsmath}
\usepackage{hyperref}
\usepackage{blindtext}
\usepackage[utf8]{inputenc}
\usepackage{graphicx} 
\usepackage{float}
\usepackage{subcaption}
\usepackage{booktabs,subcaption,amsfonts,dcolumn}
\usepackage{xspace}

\newcommand{\ourCD}[0]{GCD\xspace}
\usepackage{verbatim} 

\newcommand\Tstrut{\rule{0pt}{2.0ex}}         
\newcommand\Bstrut{\rule[-0.9ex]{0pt}{0pt}}   
\definecolor{applegreen}{rgb}{0.55, 0.71, 0.0}
\definecolor{PineGreen}{rgb}{0.0, 0.47, 0.44}
\definecolor{persiangreen}{rgb}{0.0, 0.65, 0.58}
\title{Biases, syntax and semantic in LSTM language models}
\title{Bias, bias or bias: Decomposition of LSTM language models}

\title{Analysing Neural Language Models: Contextual Decomposition Reveals Default Reasoning in Number and Gender Assignment}

\author{Jaap Jumelet \\
{\normalsize\tt jumeletjaap@gmail.com} \\
{\normalsize University of Amsterdam}\And
Willem Zuidema \\
{\normalsize\tt w.h.zuidema@uva.nl} \\
{\normalsize ILLC, University of Amsterdam}\And
Dieuwke Hupkes \\
{\normalsize\tt d.hupkes@uva.nl} \\
{\normalsize ILLC, University of Amsterdam}
}

\date{}

\begin{document}
\maketitle

\begin{abstract}
Extensive research has recently shown that recurrent neural language models are able to process a wide range of grammatical phenomena.
\textit{How} these models are able to perform these remarkable feats so well, however, is still an open question.
To gain more insight into what information LSTMs base their decisions on, we propose a \textit{generalisation} of \textit{Contextual Decomposition} (GCD).
In particular, this setup enables us to accurately distil which part of a prediction stems from semantic heuristics, which part truly emanates from syntactic cues and which part arise from the model biases themselves instead.
We investigate this technique on tasks pertaining to syntactic agreement and co-reference resolution
and discover that the model strongly relies on a \textit{default reasoning} effect to perform these tasks.
\end{abstract}

\section{Introduction}

Modern language models that use deep learning architectures such as LSTMs, bi-LSTMs and Transformers, have shown enormous gains in performance in the last few years and are finding applications in novel domains, ranging from speech recognition and writing assistance to autonomous generation of fake news.
Understanding how they reach their predictions has become a key question for NLP, not only for purely scientific, but also for practical and ethical reasons. 

From a linguistic perspective, a natural approach is to test the extent to which these models have learned classical linguistic constructs, such as inflectional morphology, constituency structure, agreement between verb and subject, filler-gap dependencies, negative polarity or reflexive anaphora. An influential paper using this approach was presented by \citet{Linzen2016AssessingDependencies}, who investigated the performance of an LSTM-based language model on number agreement. In many later papers \citep[e.g.][]{Gulordava2018ColorlessHierarchically, Wilcox2018WhatDependencies,Jumelet2018DoItems,Marvin2018TargetedModels,Giulianelli2018UnderInformation} 
a wide spectrum of grammatical phenomena has been investigated, assessing these grammatical abilities in a mainly ``behavioural'' fashion, by considering the model's output. 

In this paper, we take it as established that neural language models have indeed learned a great number of non-trivial linguistic patterns and ask instead \textit{how} language models come to show this behaviour, and, more specifically, what kind of information they use to come to their decisions. 
There exist already a number of approaches that look inside the high-dimensional vector representations and non-linear functions of these models, trying to track the flow of information. 
In the next section, we will review some of that work, distinguishing between hypothesis-driven and data-driven methods.
We highlight in particular one method called Contextual Decomposition \citep[CD, ][]{Murdoch2018BeyondLSTMs}, that combines the strengths of hypothesis- and data-driven analysis methods.

In the remainder of this paper, we then propose a generalisation of this method, which we call Generalised Contextual Decomposition (``\ourCD''). 
We derive equations for \ourCD for the case of a unidirectional (one or multi-layer) LSTM \cite{Hochreiter1997LongMemory}, and use the method to analyse how a language model processes two different phenomena: number agreement and gendered pronoun resolution. 

We demonstrate the power of \ourCD through the revelation of some important asymmetries in the way that both the singular-plural and the male-female distinction are handled.
In particular, we find evidence for a \emph{default reasoning} effect, which we believe could also be important for future work on detecting and removing bias: a default category (singular, masculine) appears to be hard-coded in the weights of the language model, number and gender information in the word embeddings themselves mainly plays a role for phrases of the opposite category (plural, feminine).
Furthermore, \ourCD enables us to investigate pronoun resolution in a way that has not been done before: by delving into the model reasoning we are able to accurately pinpoint where and how this resolution takes place.\footnote{
We have integrated all our code in \href{https://github.com/i-machine-think/diagnnose}{\tt diagnnose} \cite{diagnnose}, a well-documented analysis library which facilitate the diagnosis of neural network activations: \href{https://github.com/i-machine-think/diagnnose}{\tt\scriptsize github.com/i-machine-think/diagnnose}.
}

\section{Network analysis methods}

Recently, methods to open the blackbox of deep neural networks have become an important research area \citep[see][for recent reviews of proposed methods in NLP]{poernerRothSchutze18acl,belinkov2019analysis}. 
We distinguish between hypothesis-driven methods, and data-driven methods.
Hypothesis-driven methods include 
probes or \emph{diagnostic classifiers}, that test whether specific, a priori defined information can be decoded from the internal states of a neural model,  
many ablation studies, and types of correlation analysis, where correlations between the structure of internal representations of better and lesser understood models are studied).
An example of this approach is \citet{Giulianelli2018UnderInformation}, who trained linear diagnostic classifiers on all layers and gate activations of an LSTM to predict the number of the subject that the verb, occurring later in the sentence, needs to agree with (i.e.\ the number-agreement task). 
Their results show that the relevant information is encoded in a different way in different components of the model, and at different times while processing a sentence. 
This result is interesting, because it starts from a clearly interpretable hypothesis (number information must be maintained somewhere while the network traverses the sentence), but the work also demonstrates the limitations of the approach: It progresses one hypothesis about one linguistic pattern at a time and involves much training, work, and computation at each step.

Data-driven methods include gradient-based methods and contextual decomposition. 
An example of a gradient-based method is \citet{arras2017explaining}, who adapt Layer-wise Relevance Propagation \cite[LRP,][]{bach2015pixel} to the case of LSTMs. The key idea is to run the LSTM on each input of interest (the forward pass), then define a relevance vector at the output layer and propagate that relevance backwards through the network. 
The relevance vector simply singles out the dimensions of the output of interest, and sets all other dimension to zero. 
The backward pass is almost standard backpropagation, except that relevance does not backpropagate into the gates.
While \citeauthor{arras2017explaining}'s results reveal interesting patterns in sentences used in a sentiment classification task, their work illustrates some limitations as well. 
In particular, the work deals with a classification task with few classes, aggregates relevance per word for each predicted class, but offers little insight in how word meanings interact to build up sentence meaning beyond `pushing in the right direction' vs. `pushing in the wrong direction'.

An alternative data driven method, and the one that we will expand on in this paper, is Contextual Decomposition for LSTMs \cite[CD,][]{Murdoch2018BeyondLSTMs}. 
The key idea behind this technique is to partition the hidden states into two components, that \citeauthor{Murdoch2018BeyondLSTMs} label `relevant' and `irrelevant'. 
For each word in a sentence, they do a forward pass that computes all cell and gate activations as in normal operation of the neural network, but also partition each activation value of each neuron in $h$ or $c$ in a part that is \emph{caused} by some selected token or phrase in focus, and a part that is not. 
They achieve this by deriving a factorisation of the update formulas for $h$ and $c$, that expresses them as a long sum of components and then selecting some of these components as being relevant, and others as irrelevant. 
Qualitative results on sentiment analysis suggest that CD can attribute roles to words in a sentence very well, better than alternatives the authors considered (which, unfortunately, did not include LRP).

CD thus requires no extra training and requires only the forward pass of the network.
It can easily be extended to work efficiently with many classes, such as the language modelling task that we are interested in. 
In the next section, we will define CD more precisely, where we will use the terms \textit{inside} and \textit{outside} rather than relevant and irrelevant. 
We then propose a generalisation that allows us to experiment with different \emph{hypotheses} on what goes into the \textit{inside} and \textit{outside} bins, enabling some of the advantages of hypothesis-driven analysis methods to be brought into this data-driven method.

\begin{figure*}[t]
    \centering
    \includegraphics[width=0.85\textwidth]{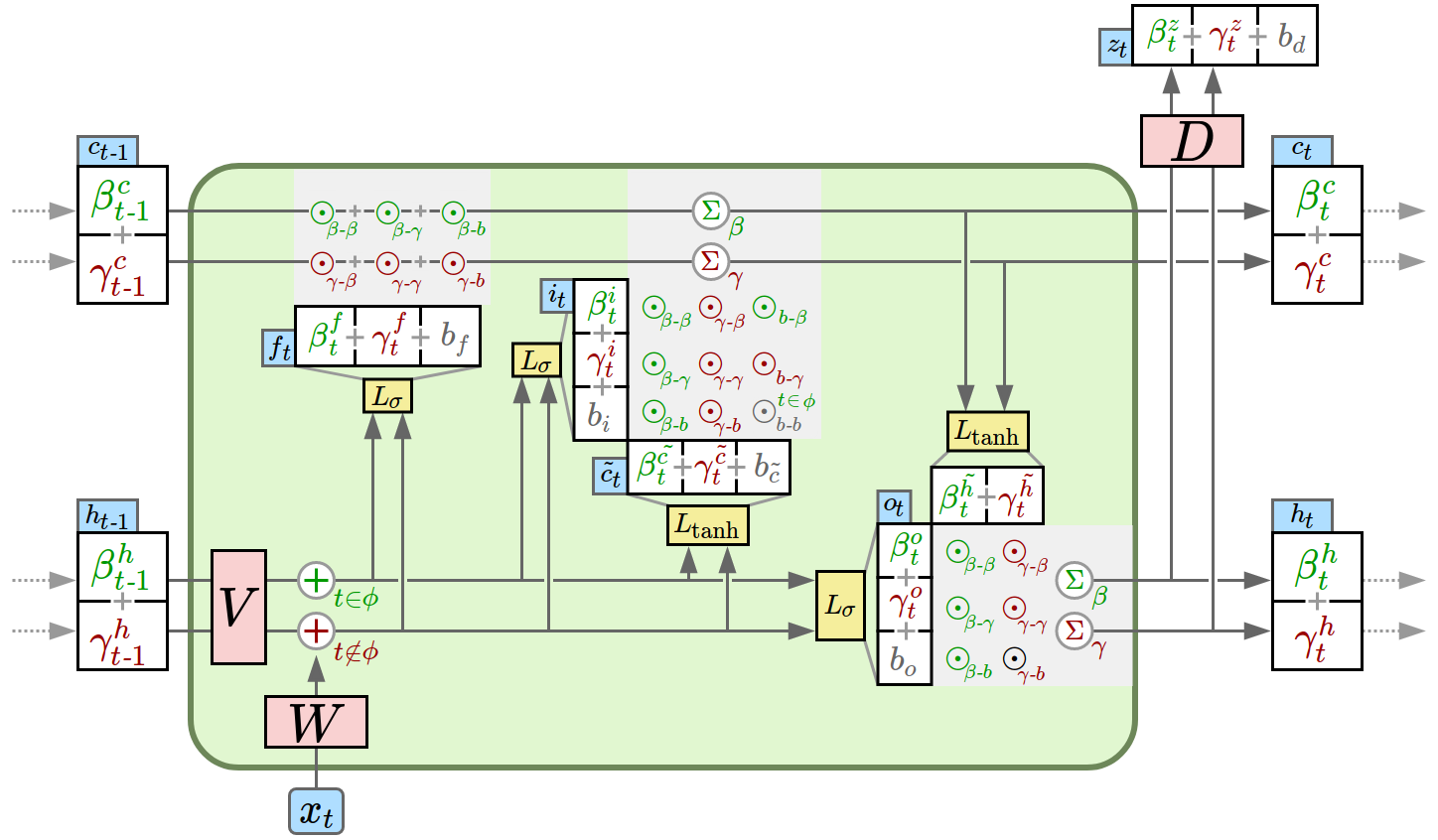}
\caption{
A graphical overview of \ourCD, based on the LSTM design of \citet{colah2015}. 
$\phi$ denotes the phrase in focus, and $t\in\phi$ implies the action is only performed when step $t$ is part of $\phi$.
$\odot$ denotes an individual interaction; green interactions are added the $\beta$ part and red interactions to $\gamma$. 
$V$, $W$, and $D$ represent the linear projections of the LSTM itself.
The interaction set denoted here corresponds to the \textsc{in} set of Equation \ref{eq:our_rel}. 
}\label{fig:cd-overview}
\end{figure*}
\section{Generalised Contextual Decomposition}\label{methods}
In this particular study, we consider the LSTM language model that was made available by \citet{Gulordava2018ColorlessHierarchically}. 
This language model (LM) is a 2-layer LSTM with 650 hidden units in both layers, trained on a corpus with Wikipedia data.
Given the relevance of the specific LSTM-dynamics for the understanding of the main method of our paper, we repeat the equations that describe it below.

\vspace{-2mm}
\setlength{\abovedisplayskip}{0pt}
\setlength{\belowdisplayskip}{5pt}
\setlength{\abovedisplayshortskip}{0pt}
\setlength{\belowdisplayshortskip}{0pt}
{\small\begin{align}
    f_t &= \sigma(W_fx_t + V_fh_{t-1} + b_f)\label{eq:forget}\\
    i_t &= \sigma(W_ix_t + V_ih_{t-1} + b_i)\label{eq:input}\\
    \tilde{c}_t &= \tanh(W_{\tilde{c}}x_t + V_{\tilde{c}}h_{t-1} + b_{\tilde{c}})\label{eq:cell}\\
    o_t &= \sigma(W_ox_t + V_oh_{t-1} + b_o)\label{eq:output}\\[2pt]
    \label{gates}c_t &= f_t \odot c_{t-1} + i_t \odot \tilde{c}_t\\\label{eq:hidden}
    h_t &= o_t \odot \tanh(c_t)\\[2pt]
    z_t &= Dh_t + b_d\\
    p_t &= \textup{SoftMax}\left(z_t\right)
\end{align}
}%
The final model output $p_t$ represents a multinomial distribution over the model's vocabulary.
Throughout the paper we refer to the bias terms $b$ as the model \textit{intercepts}, to avoid confusion with general biases that the model may have.

\paragraph{CD} To compute the contributions of one or multiple input tokens (said to be \textit{in focus}) to the output of an LSTM cell, \citet{Murdoch2018BeyondLSTMs} divide each cell and hidden state into a sum of two parts: a $\beta$ part, which contains the part of this particular state that stems from \textbf{inside} this phrase, and a $\gamma$ part, which contains information coming from words \textbf{outside} this phrase.
The output logit $z_t$ can then be redefined as

\vspace{-2mm}{\small\begin{align*}z_t=Dh_t+b_d&= D\beta^h_t+D\gamma^h_t+b_d\\&=\beta^z_t+\gamma^z_t+b_d\end{align*}}with $\beta^z_t$ providing a quantitative score of the phrase's contribution to the logit.
How a particular hidden state $h_t$ is partitioned into $\beta^h_t$ and $\gamma^h_t$ 
is determined by two things: \textbf{i)} The decomposition of the \textit{previous} states $c_{t-1}$ ($\beta^c_{t-1}$ and $\gamma^c_{t-1}$) and $h_{t-1}$ ($\beta^h_{t-1}$ and $\gamma^h_{t-1}$), and \textbf{ii)} 
Which \textit{interactions} between the different $\beta$ and $\gamma$ terms, the intercepts $b$, and the input $x_t$ are considered to be part of the inside contribution of the phrase. 
We provide a graphical overview of our setup in Figure \ref{fig:cd-overview}.

\paragraph{Factorised activation functions}
The gate interactions cannot yet be expanded into a cross-term of their input parts, due to the non-linear activation that wraps them.
\citeauthor{Murdoch2018BeyondLSTMs} define a method to factorise the sigmoid and $\tanh$ functions for each specific gate into a sum of contributions of the input terms, such that
\vspace{2mm}\[\tanh(\sum\nolimits_{i=1}^Ny_i) = \sum\nolimits_{i=1}^NL_{\tanh}(y_i)\]
$L_{\tanh}$ expresses the \textit{contribution} of each input, which is computed by averaging over the differences of all possible permutations of the input terms; a procedure that corresponds to the calculation of the Shapley values \cite{shapley1953value}.\footnote{In the original formulation this procedure is called \textit{linearizing}. We deemed this term to be slightly confusing, as the resulting functions $L$ are still non-linear.}

Before this factorisation is performed, an input token $x_t$ is added to the inside part $\beta$ if it is part of the phrase for which we decompose (i.e.\ the phrase \textit{in focus}), otherwise it is added to $\gamma$.
Equation~\ref{eq:forget}, for example, can then be rewritten as:

\vspace{-1mm}
{\small
\begin{align}
    \begin{split}
        f_t &= \sigma\left(V_f\beta^h_{t-1} + V_f\gamma^h_{t-1} + W_fx_t + b_f \right)\label{eq:linforget}\\
        &= L_{\sigma}\left(V_f\beta^h_{t-1}\textcolor{persiangreen}{+W_fx_t}\right) + L_{\sigma}(\gamma^h_{t-1} ) + L_{\sigma}(b_f)
    \end{split}
\end{align}
}%
where $x_t$ is considered to be inside the phrase in focus and therefore added to the $\beta$ part (denoted in \textcolor{persiangreen}{green} for extra emphasis).
A similar sum can be written down for the input gate $i_t$ and the candidate cell state $\tilde{c}_t$.
This allows the two products $f_t\odot c_t$ and $i_t\odot\tilde{c}_t$ of Equation~\ref{gates} to be expanded into a sum of cross-terms between the decomposed gate and (candidate) cell values.
Expanding the forget and input gate results in 15 cross-terms, that each express different interactions between the current input, previous $\beta$ and $\gamma$ terms, and the model intercepts. 

\citeauthor{Murdoch2018BeyondLSTMs} state they observed improvements when the intercept term is fixed to the first position in each permutation.
Consequently, however, these intercepts are assigned a relatively larger contribution, as their fixed position makes their contribution independent of the magnitudes of the other terms.
We therefore pose that the full set of permutations should be considered, to assign unbiased contributions to each input term.\footnote{We only discovered the impact of this decision after the paper had already been reviewed. While using the full set of permutations did, fortunately, not qualitatively change our conclusions,  the exact numbers presented in this work thus differ from the earlier version of this paper. For completeness, we report the original results with the fixed intercept positions in the supplementary materials of this article.}

\paragraph{Decomposing interactions}
Based on all the different interaction terms, the decomposition is determined by \textit{which} of these interactions should be considered to belong to the inside part $\beta$ of the next cell state and which to the outside part $\gamma$.

In the formulation of \citeauthor{Murdoch2018BeyondLSTMs}, all interactions with outside parts $\gamma_t$ are disregarded for the computation of $\beta_{t+1}$, and therefore only information directly stemming from the $\beta_t$ terms with no interference from $\gamma_t$ is taken into account.
Of the 15 cross-product terms described above, this leaves  5 terms to be part of $\beta^c_{t+1}$: 

{\small
\begin{align}\label{eq:default_rel}
    \beta^c_{t+1} &= L_{\sigma}(V_f\beta^h_t\textcolor{persiangreen}{+W_fx_t})\odot\beta^c_t && \beta\textup{-}\beta\nonumber\\ 
      &+ L_{\sigma}(b_f)\odot\beta^c_t&& \beta\textup{-}b\nonumber\\
      &+ L_{\sigma}(V_i\beta^h_t\textcolor{persiangreen}{+W_ix_t})\odot L_{\tanh}(V_{\tilde{c}}\beta^h_t\textcolor{persiangreen}{+W_{\tilde{c}}x_t})&& \beta\textup{-}\beta\nonumber\\
      &+ L_{\sigma}(V_i\beta^h_t\textcolor{persiangreen}{+W_ix_t})\odot L_{\tanh}(b_{\tilde{c}}) && \beta\textup{-}b\nonumber\\
      &+ L_{\tanh}(V_{\tilde{c}}\beta^h_t\textcolor{persiangreen}{+W_{\tilde{c}}x_t})\odot L_{\sigma}(b_i)&& \beta\textup{-}b
\end{align}
}%
The remaining 10 terms from the cross-product are put in $\gamma^c_{t+1}$. 
We use the notation $\{\beta$-$\beta$, $\beta$-$b\}$ to concisely describe this set of interactions. 
The decomposition of the hidden state is created by decomposing the output gate:

{\small\begin{align}\label{eq:dec_hidden}
\begin{split}
    \beta^h_{t+1} &= L_{\sigma}(V_o\beta^h_t\textcolor{persiangreen}{+W_ox_t})\odot\beta^c_{t+1}\\ 
      &+ L_{\sigma}(b_o)\odot\beta^c_{t+1}
\end{split}
\end{align}
}%
The decomposed contribution score $\beta^z_T$ over the model vocabulary at step $T$ of some phrase in focus is then calculated by passing the decomposed hidden state to the decoder, i.e.~$D\beta^h_T$. 
This score can be expressed as a relative contribution by normalising it by the full model logit $z$ (including $b_d$).
In a multi-layer LSTM, $\beta$ and $\gamma$ parts are not only propagated \textit{forward}, but also \textit{upward}, where they are added to their respective parts in the layer above them.
For initialisation $\beta$ is set to a zero vector, and $\gamma$ is set to the initial LSTM states.\footnote{For the initial states we use the activations that follow from the short phrase ``\texttt{.} \texttt{<eos>}''. This phrase resets the model state to a clean slate, and leads to better results than using 0-valued activations.}

\paragraph{Generalising CD}
\label{subsec:CD_for_LM}
\label{subsec:interactions}

While \citet{Murdoch2018BeyondLSTMs} consider only one way of partitioning interactions between inside and outside components, their setup can be quite easily generalised to also allow other interactions to be included in the inside terms $\beta$.
To obtain a better insight into how different interactions contribute to the final prediction, we experiment with various ways of defining the set of relevant interactions.

A particular case concerns the interactions between $\beta$ and $\gamma$. 
It wouldn't be correct to completely attribute the information flowing from these interactions to the phrase in focus, but disallowing any information stemming from interactions of a phrase with a subsequent token results in loss of relevant information.
Consider, for instance, the verb prediction in a number agreement task. 
While the correct verb \emph{form} depends only on the subject, the right \emph{time} for this information to surface depends on the material in between, which in the setup described in Equation~\ref{eq:default_rel} would be discarded by assigning the $\beta$-$\gamma$ interactions to $\gamma$.

Taking inspiration from \citet{arras2017explaining}, and based on their motivation,
we add the $\beta_s$-$\gamma_g$ interaction to the relevant interaction set, while still disregarding a $\gamma_s$-$\beta_g$ interaction.
The $g$ subscript denotes the part of the interaction that is coming from the gate, and $s$ the source part.
We denote this amended interaction as $\beta$-$\gamma^*$.

Furthermore, we follow the addition of \citet{Singh2019HierarchicalPredictions} of only adding the intercept interactions $b$-$b$ to the inside part if the current time step is part of the phrase in focus, which we denote as \mbox{$b$-$b \in x$}.
We add these $\beta$-$\gamma^*$ and $b$-$b \in x$ interactions to Equation \ref{eq:default_rel}, resulting in the following decomposition that is presumed to come from \textbf{inside} the phrase in focus (denoted as \textsc{in}):

\vspace{-2mm}{\small
\begin{align}\label{eq:our_rel}
\beta^c_{t+1} &= \{\beta\textup{-}\beta, \beta\textup{-}b\} &\nonumber\\
  &+ L_\sigma(V_f\gamma^h_t)\odot \beta^c_t  && \beta\textup{-}\gamma^*\nonumber\\
  &+ L_\sigma(V_i\gamma^h_t)\odot L_{\tanh}(V_{\tilde{c}}\beta^h_t\textcolor{persiangreen}{+W_{\tilde{c}}x_t})  && \beta\textup{-}\gamma^*\nonumber\\
  &+\textcolor{persiangreen}{L_\sigma(b_i)\odot L_{\tanh}(b_{\tilde{c}})} && b\textup{-}b\in x
\end{align}
}%

We also experimented with various other interaction sets.
To determine the influence of the gate intercepts, we create an interaction set that does not take the input embeddings into account at all: $\{\beta$-$\beta, \beta$-$\gamma^*, \beta$-$b, b$-$b\}$, with $x$ always added to $\gamma$, denoted as \textsc{intercept}$^*$.
We include $\beta$-$\gamma^*$ to still account for the way the intercepts are gated by the input sentence. 
The initial hidden and cell state are added to $\beta$ now as well, as we consider these states to be part of the model bias.
Finally, to determine the dependence of the input on the gate intercepts we use an interaction set that never takes the interactions with any intercept into account: $\{\beta$-$\beta, \beta$-$\gamma^*\}$, denoted as $\neg$\textsc{intercept}.

\begin{figure*}[t]
    \centering
    \begin{subfigure}{0.32\linewidth}
        \centering
        \includegraphics[width=1.15\textwidth, trim=0 0 0mm 0, clip]{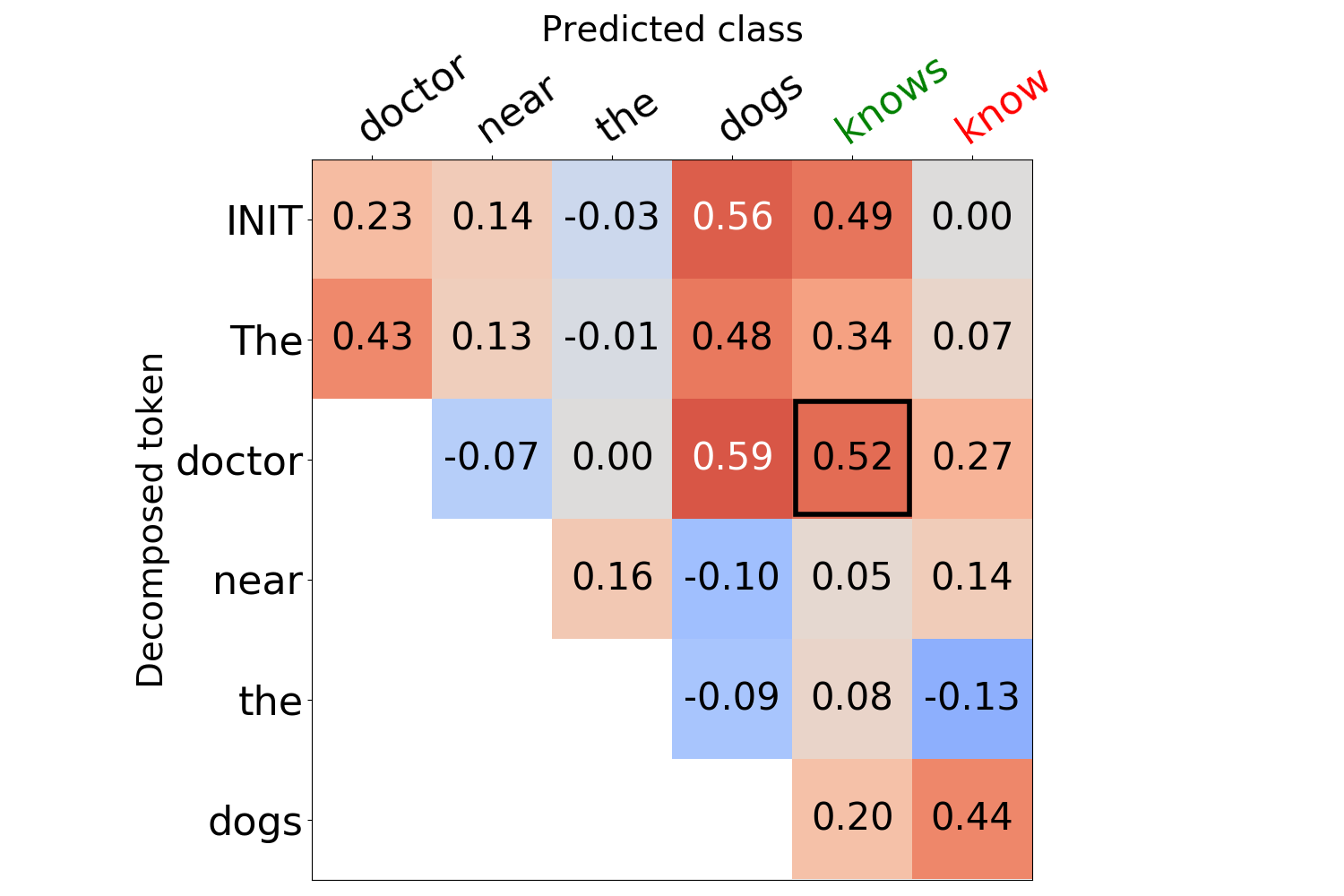}
        \caption{A single \textit{NounPP} sentence: singular subject with plural attractor (SP).}\label{fig:attention_sp}
    \end{subfigure}
    \hfill
    \begin{subfigure}{0.32\linewidth}
        \centering
        \includegraphics[width=1.15\textwidth, trim=0 0 0mm 0, clip]{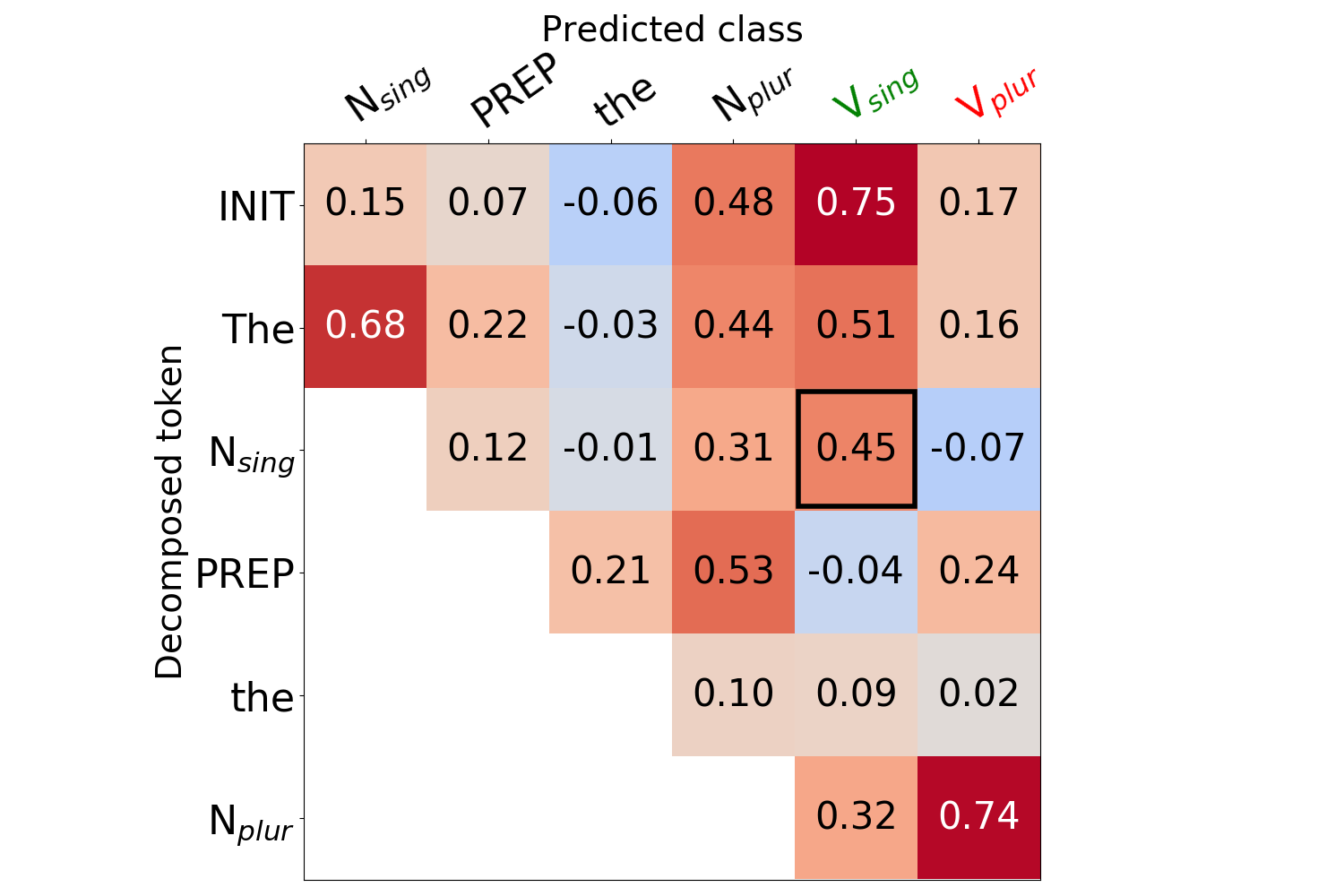}
        \caption{Average \textit{NounPP} SP: singular subject with plural attractor.}\label{fig:attention_sp}
    \end{subfigure}
    \hfill
    \begin{subfigure}{0.32\linewidth}
        \centering
        \includegraphics[width=1.15\textwidth, trim=0 0 0mm 0, clip]{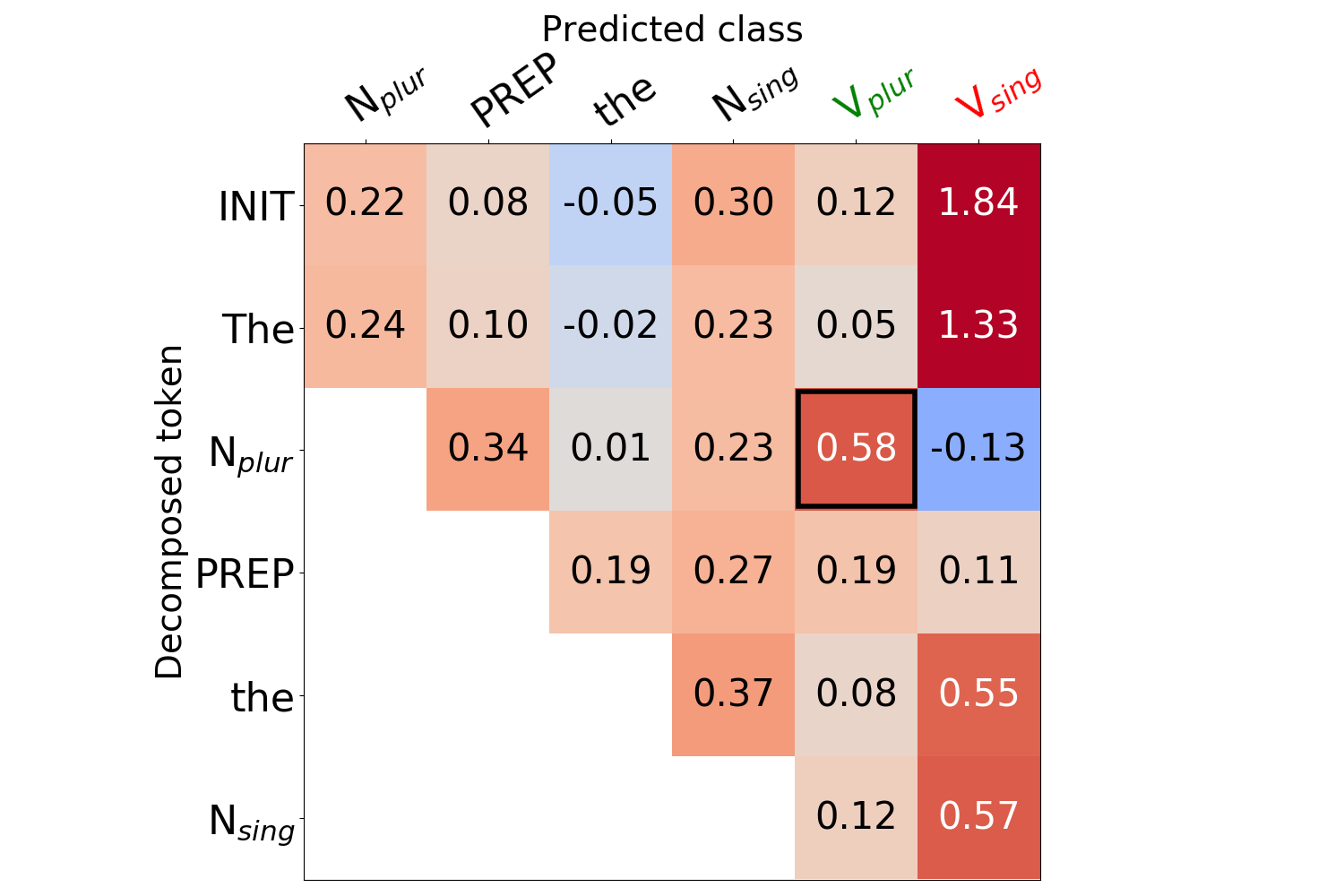}
        \caption{Average \textit{NounPP} PS: plural subject with singular attractor.}\label{fig:attention_ps}
    \end{subfigure}
    \caption{Average contributions for the NounPP corpus of \citet{Lakretz2019TheModels}, defined as $\beta^z_t/z_t$. \texttt{INIT} denotes the contribution of the initial states. The picture depicts an asymmetry in the way that the model encodes singularity and plurality: while plural verbs depend strongly on the subject, for singular sentences this is not the case. }\label{fig:sv_cd_matrix}
\end{figure*}
\section{Experimental setup}

We use \ourCD to study how our LSTM model handles two different linguistic phenomena: subject-verb agreement and anaphora resolution in relation to \emph{gender}.
Next to the model of \citeauthor{Gulordava2018ColorlessHierarchically} (for which we present our results), we also ran our experiments on the LM of \citet{googlelm}, which arrives at similar results.

\subsection{Subject-verb agreement}
We consider a variant of the \textit{number-agreement} (NA) task that was proposed by \citet{Linzen2016AssessingDependencies} to assess the syntax-sensitivity of language models.
In this task, a model is evaluated based on its ability to track a long-distance subject-verb relation, which is assessed by the percentage of times that the verb-form it prefers matches the \textit{number} of the syntactic subject.
Commonly, the material in between subject and verb contains an \textit{attractor} noun that competes with the syntactic subject, e.g.\ \textit{The \textbf{keys} on the \underline{table} \textbf{are}}.

Here, we consider the NA corpora made available by \citet{Lakretz2019TheModels}, which consists of a number of data sets containing a range of syntactic constructions in which number agreement plays a role. 
We report results for several of their data sets, but focus in particular on their \emph{NounPP} subset, in which sentences contain an attractor embedded in a prepositional phrase.
These sentences are formed following the template \emph{The} \texttt{N} \texttt{Prep} \emph{the} \texttt{N} \texttt{V} [..], e.g.\ \emph{The boys near the car greet} [..].
The sentences in this data set are split based on the number of the subject and the attractor, resulting in four different conditions: SS, SP, PS, and PP.

\subsection{Anaphora resolution and gender bias}\label{ref:data_gender}
Our second experiment concerns anaphora resolution and the possible gender biases that networks may use to perform this task.
We focus on intra-sentential anaphora resolution, in which a pronoun in a subordinate clause refers to an entity in the main clause, based on \mbox{\textit{gender information}}. 
For example: \textit{The \underline{monk} liked the \textbf{nun}, because \textbf{she} was always nice to \underline{him}}.

Compared to number agreement it is more difficult to formulate a setup for anaphora resolution in which there is a right or wrong prediction that directly reflects how the model handles the phenomenon: when predicting \textbf{\textit{she}} in the example, it could have been equally probable to predict \textbf{\textit{he}}.
Rather, to establish if a model correctly resolves the referent of a pronoun, it should be checked what the model considered to be the source of this prediction, which cannot directly be inferred from the prediction itself.
\ourCD gives us exactly this information and is therefore an excellent tool to study anaphora resolution in language modelling.

To create our corpus, we use the templates from the WinoBias corpus created by \citet{zhao2018gender}.
This corpus contains sentences with job titles that are gender neutral, yet contain a stereotypical bias towards one gender (doctors and CEOs are \textit{male}, nurses and housekeepers \textit{female}).
We construct two types of corpora, one containing the stereotypical job titles of \citeauthor{zhao2018gender} and one in which we replace these titles by entity descriptions that are unambiguously gendered (\textit{king}, \textit{bride}, \textit{father}, etc.). 
Similar to the \textit{NounPP} corpus for NA, we create 4 different conditions, based on the gender of the subject and object (FF, FM, MF, and MM).
An example of an MF sentence would be \textit{The father likes the woman, because he/she}. 
We sample from the set of entity descriptions to create 500 sentences per condition, for both corpus types.

\subsection{Experiment types}

\paragraph{Phrase contributions}
In the first type of experiment, we consider the contributions of different words in the input to a later prediction of the model.
This allows us to compare the contributions of different words in the sentence and track which words the model uses to come to its prediction.
We compute a phrase's contribution to a prediction at step $t$ as $\beta^z_t/z_t$.

\paragraph{Pruning information}
In the second type of experiment, we focus on the model's \emph{predictions}.
In particular, we study how the model's predictions change when it is forced to consider only specific parts of the input, by disregarding all information that does not belong to the inside information of that part of the input.
This allows us to quantify the extent to which a correct prediction does in fact stem from that phrase.
For this experiment, we consider several different interaction sets, that differ in what is considered to be inside the contribution of the phrase: \textsc{in} describes the direct contribution of some phrase, \textsc{intercept} the contribution of the model intercepts, and $\neg$\textsc{intercept} the contribution of some phrase without its intercept interactions.

\section{Subject-verb Agreement}
We now study what information the LM uses to achieve the high prediction accuracies that were reported by \citet{Lakretz2019TheModels}.

\subsection{Phrase contributions}
For every word in a sentence, we compute the \ourCD contribution for all words preceding this word.
We plot these contributions in a \textit{decomposition matrix} (akin to the attention plots often seen in machine translation papers).
Every cell of this matrix represents the contribution of an input $x_i$ (row i) to an output $y_j$ (column j).
The complete decomposition of an output word $y_j$ can thus be found in column $j$.
The reported scores are the decomposed scores normalised by the total model logit, resulting in the relative contribution.

In Figure~\ref{fig:sv_cd_matrix}, we plot the average decomposition matrices for the SP and PS splits of the NounPP data set. 
While many interesting observations can be made here, we would like to focus on the final 2 columns that represent the decompositions of the correct and wrong verb in the sentence, and on the contribution of the subject to this verb.
In the singular case (\ref{fig:attention_sp}), this contribution is, surprisingly, relatively low: The correct verb prediction does not seem to depend solely on the syntactic subject, but stems from elements that lie outside the subject as well.
For the plural case, this picture is strikingly different: The highest contribution now stems from the subject of the sentence.
When considering the decomposition of the wrong verb (the final column) it becomes even more clear that contributions to a plural verb predominantly stem from a plural noun, whereas singular verbs receive strong contributions from non-numbered tokens as well.
This quite remarkable difference provides the first evidence for one of our conclusions: A singular prediction acts as the default number for the model, and predicting a plural verb requires some explicit evidence coming from the subject.

\subsection{Pruning information}
To quantify to which extent the model bases its prediction on the subject, we prune all information that is not directly related to the subject and repeat \citeauthor{Lakretz2019TheModels}'s NA tasks.
If the model prediction were based solely on the number of the subject, its accuracy should go up, as we filter out all potentially intervening or confusing information.
If, on the other hand, the prediction of the verb is not causally linked to the subject, but the model is using heuristics that require the rest of the sentence, no increase in accuracy is to be expected.
We show the results, along with the accuracy of the full model in Table~\ref{tab:sv-accuracies}.

\begin{table}[t]
\footnotesize
\centering
\setlength{\tabcolsep}{2pt}
\begin{tabular}{|c|c||c|ccc|}
\cline{4-6}\multicolumn{3}{c}{} & \multicolumn{3}{|c|}{\Tstrut \textsc{gcd}} \\
\hline\textbf{Task} & \textbf{C} & \textsc{full} & \textsc{in} & \textsc{intercept}$^*$ & $\neg$\textsc{intercept}\Tstrut\\\hline\Tstrut
\textcolor{persiangreen}{Simple}    & \textcolor{persiangreen}{S}  & \textcolor{persiangreen}{100} & \textcolor{persiangreen}{73.3} \scriptsize{(91.3)} & \textcolor{persiangreen}{97.3} \scriptsize{(100)} & \textcolor{persiangreen}{69.7} \scriptsize{(86.3)} \Tstrut\\
Simple             & P           & 100        & 100  \scriptsize{(100)}          & 32.7 \scriptsize{(7.7)}         & 100 \scriptsize{(100)}  \Bstrut\\\hline\Tstrut
\textcolor{persiangreen}{nounPP}    & \textcolor{persiangreen}{SS} & \textcolor{persiangreen}{99.2} & \textcolor{persiangreen}{93.0} \scriptsize{(99.7)}& \textcolor{persiangreen}{99.8} \scriptsize{(99.8)} & \textcolor{persiangreen}{72.7} \scriptsize{(88.7)} \\
\textcolor{persiangreen}{nounPP}    & \textcolor{persiangreen}{SP} & \textcolor{persiangreen}{87.2} & \textcolor{persiangreen}{90.3} \scriptsize{(99.3)} & \textcolor{persiangreen}{98.8} \scriptsize{(99.8)} & \textcolor{persiangreen}{60.5} \scriptsize{(83.5)} \\
nounPP             & PS          & 92.0          & 100   \scriptsize{(100)}        & 0.0   \scriptsize{(0.0)}         & 100 \scriptsize{(100)} \\
nounPP             & PP          & 99.0          & 100   \scriptsize{(99.3)}        & 7.0 \scriptsize{(0.5)}         & 99.8  \scriptsize{(100)} \Bstrut\\\hline\Tstrut
\textcolor{persiangreen}{namePP}    & \textcolor{persiangreen}{SS} & \textcolor{persiangreen}{99.3} & \textcolor{persiangreen}{97.7} \scriptsize{(91.3)} & \textcolor{persiangreen}{99.4} \scriptsize{(100)} & \textcolor{persiangreen}{76.2} \scriptsize{(90.9)} \\
namePP             & PS          & 68.9          & 98.3 \scriptsize{(98.2)}          & 1.3 \scriptsize{(0.0)}            & 99.9 \scriptsize{(99.9)}  \Bstrut\\\hline
\end{tabular}
\caption{
Accuracies on various subject-verb agreement tasks of \citet{Lakretz2019TheModels}. 
\textsc{full} denotes the full model accuracies. 
\textsc{in} is the decomposition of the subject,
\textsc{intercept}$^*$ only decomposes the gate intercepts of the model. 
$\neg$\textsc{intercept} takes no interactions with the intercepts into account.
Singular conditions are denoted in \textcolor{persiangreen}{green}. 
$(\cdot)$ denotes accuracies of scores without decoder bias, i.e. $Dh_t$ vs $Dh_t+b_d$.
}\label{tab:sv-accuracies}
\end{table}

These numbers show a strong causal relation between plural subjects and verbs: The number prediction accuracy for the \textsc{in} decomposition goes up for all cases with a plural subject.
This confirms our previous finding from the decomposition matrix, which showed a relatively high contribution of plural subjects to plural verbs, as well as the conclusion of \citet{Lakretz2019TheModels} that the model is in fact keeping track of syntactic structure.

When considering the singular subjects an interesting pattern emerges: The decomposition of sentences for which the intervening attractor has the same number leads to a \textit{lower} accuracy.
This confirms that the model is in fact basing its prediction for these conditions on information that lies outside the subject itself.

\paragraph{Intercepts}
When we only decompose with respect to the gate intercepts (\textsc{intercept}$^*$, column 5) it turns out the model has an extreme preference for selecting singular verbs.
Decomposing without the intercept interactions (\textsc{no intercept}, column 6) leads (as expected) to opposite results: the decomposed model now has a strong preference towards plural verbs as the singular prediction no longer can depend on these intercepts.
This further confirms that singular verbs are used as a default baseline, which is partly encoded in its intercepts.
To predict plural verbs, on the other hand, some evidence is needed, which the model picks up correctly from the subject number.

\paragraph{Corpus frequency}
One would expect that due to the model's default number being singular, this class to be more encountered during training.
This turns out not to be the case: in the model's training corpus the plural verbs of the NA tasks occurred over 5 times as often as their singular counterparts.
This higher frequency is in fact represented in the decoder intercept, which is higher on average for plural verbs, but it is surprising that the LSTM weights encode a default for the minority class.

\section{Anaphora-resolution and gender}

For the NA-tasks, the full model accuracy provides evidence that the model can perform the task well; for anaphora resolution, it is not possible to create such accuracies based on the full model predictions alone.
In this section, we therefore address two different questions: 1) \textit{Does the model correctly resolve referents?} 
In other words: When the model generates a male or female pronoun, does it consistently do this based on male and female referents encountered earlier in the sentence, and 2) If the model correctly performs anaphora resolution, what types of interactions and information does it use to do so?
In our analysis we furthermore consider the difference between sentences with unambiguously gendered referents with sentences in which the gender of the referents is ambiguous but contains a stereotypical male or female bias.

\subsection{Phrase contributions}\label{sec:gender-exp-1}
As the template that the sentences in our anaphora data set follow is not as rigid as those of the NA tasks, creating an averaged decomposition matrix for all words in the sentences does not result in a comprehensive picture.
To evaluate whether the model links pronouns to referents of the correct gender, we subtract the referent contribution to \textit{she} from that to \textit{he}: ${\beta^z_{\textit{he}}}/{z_{\textit{he}}} - \beta^z_{\textit{she}}/z_{\textit{she}}$. 
A positive difference then indicates this referent had a greater contribution towards predicting \textit{he} than \textit{she}, and a negative difference vice versa.
Little difference indicates that the referent did not contribute much to the gender of the predicted pronoun.

\begin{figure}[!t]
    \begin{subfigure}{0.47\linewidth}
        \centering
        \includegraphics[width=0.9\textwidth, trim=0 0 0mm 0, clip]{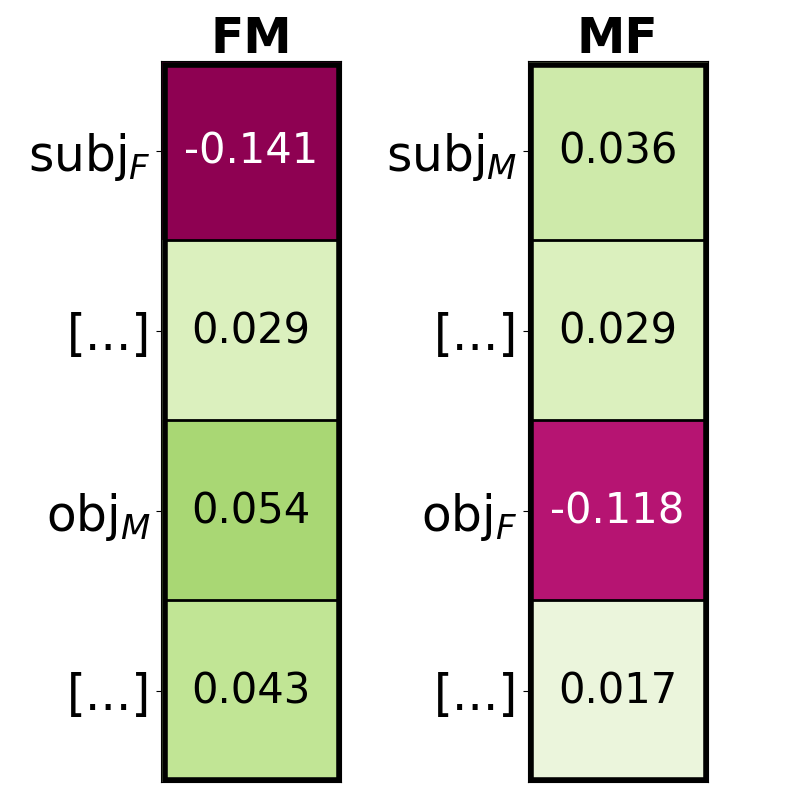}
        \caption{unambiguous}\label{fig:attention_unamb}
    \end{subfigure}
    \hfill
    \begin{subfigure}{0.47\linewidth}
        \centering
        \includegraphics[width=0.9\textwidth, trim=0 0 0mm 0, clip]{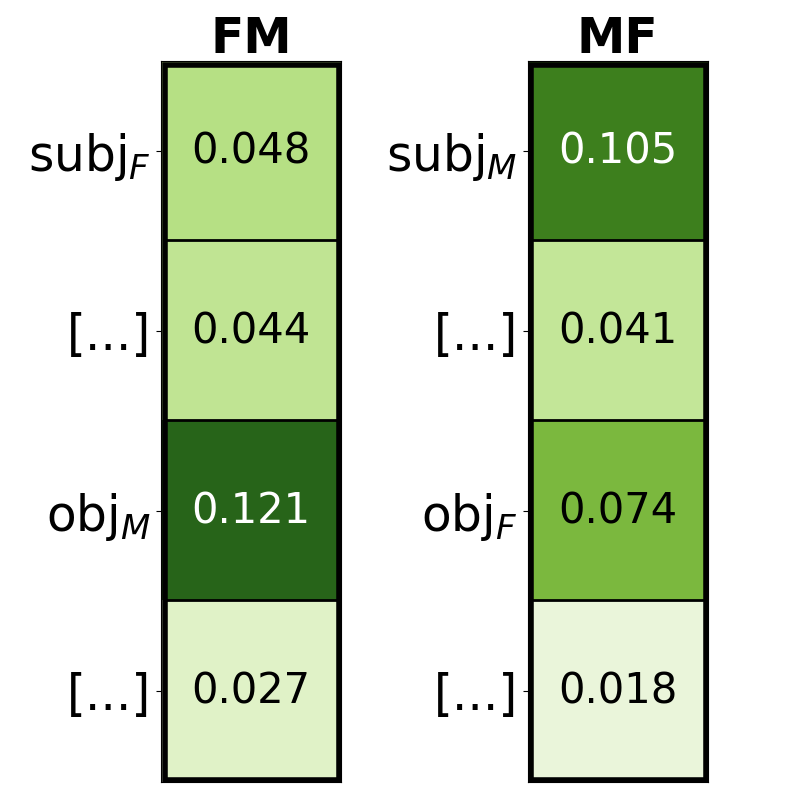}
        \caption{stereotypical}\label{fig:attention_stereo}
    \end{subfigure}
\caption{
Average decomposed preference of \textit{he} over \textit{she}, calculated as the difference between the relative contributions: ${\beta^z_{\textit{he}}}/{z_{\textit{he}}} - \beta^z_{\textit{she}}/z_{\textit{she}}$. 
Positive values denote male preference, negative values female preference.
Phrases occurring between subject and object, and object and pronoun are denoted with \texttt{[...]}.
}\label{fig:gender_attention}
\end{figure}
\begin{table*}[!ht]
\footnotesize
\centering
\begin{subtable}[t]{0.48\textwidth}
\centering
    \begin{tabular}{|c||c|ccc|}
    \cline{3-5}\multicolumn{2}{c}{} & \multicolumn{3}{|c|}{\Tstrut \textsc{gcd}} \\
        \hline
         & \textsc{full} & \textsc{subject} & \textsc{object} & \textsc{intercept}$^*$ \Tstrut\Bstrut\\
        \hline\Tstrut
        MM  & 100 & 100 \scriptsize{(93.2)} & 100 \scriptsize{(97.8)} & 100 \scriptsize{(93.2)} \Tstrut\\
        MF  & 58.6 & 100 \scriptsize{(86.4)} & 47.2 \scriptsize{(0.8)} & 100 \scriptsize{(96.0)} \\
        FM  & 37.0 & 29.2 \scriptsize{(0.6)} & 100 \scriptsize{(97.2)} & 100 \scriptsize{(98.0)}\\
        FF  & 1.2 & 77.2 \scriptsize{(0.8)} & 88.8 \scriptsize{(1.2)} & 100 \scriptsize{(92.2)}\\
        \hline
    \end{tabular}
    \caption{\%\textit{he}\textgreater \textit{she}, unambiguous referents}\label{tab:preds_unamb}
\end{subtable}
\quad
\begin{subtable}[t]{0.48\textwidth}
\centering
    \begin{tabular}{|c||c|ccc|}
    \cline{3-5}\multicolumn{2}{c}{} & \multicolumn{3}{|c|}{\Tstrut \textsc{gcd}} \\
        \hline
         & \textsc{full} & \textsc{subject} & \textsc{object} & \textsc{intercept}$^*$ \Tstrut\Bstrut\\
        \hline\Tstrut
        MM  & 100 & 100 \scriptsize{(100)} & 100 \scriptsize{(100)} & 100 \scriptsize{(88.0)}\Tstrut\\
        MF  & 94.6 & 100 \scriptsize{(99.6)} & 95.4 \scriptsize{(84.0)} & 100 \scriptsize{(84.8)}\\
        FM  & 88.8 & 90.6 \scriptsize{(77.4)} & 100 \scriptsize{(100)} & 100 \scriptsize{(91.0)}\\
        FF  & 84.6 & 92.8 \scriptsize{(75.6)} & 97.4 \scriptsize{(84.0)} & 100 \scriptsize{(89.2)}\\
        \hline
    \end{tabular}
    \caption{\%\textit{he}\textgreater \textit{she}, stereotypical referents}\label{tab:preds_stereo}
\end{subtable}

\caption{
Gender preference on the fixed and stereotypical gender corpora. 
Reported scores are the percentage of times \textit{he} is preferred over \textit{she}. 
The first column denotes the gender of the subject and object. 
\textsc{full} denotes the full model preference, \textsc{subject} the decomposed score of the subject phrase (including determiners), and \textsc{object} the decomposed object score. 
\textsc{intercept}$^*$ is the decomposed score with relation to the intercepts only.
$(\cdot)$ denotes accuracies of scores without decoder bias, i.e. $Dh_t$ vs $Dh_t+b_d$.
}\label{tab:gender_accs}\vspace{-4mm}
\end{table*}
\paragraph{Unambiguous referents}
In Figure~\ref{fig:attention_unamb} we plot this relative contribution difference for the two conditions in our data set that contain both an unambiguous female and male referent.
It is evident that the model bases its prediction on a referent of the right gender: The female subjects and objects contribute more to the prediction of \textit{she} (reflected by the negative purple cells) and the male subjects and objects more to the prediction of \textit{he} (the positive green cells).

Interestingly, this effect is much stronger visible for the female connections.
The reason for this can be found in the model intercepts; male preference is more strongly encoded in the intercepts of the decoder: \textit{he} has an intercept of 7.75, \textit{she} only 6.09.
This enables the model to use this male prediction as a default, similar to how singular verbs acted as a default baseline for number prediction.
Akin to number agreement the model thus needs to encounter sufficient evidence of an entity being female to prefer a female pronoun.
In the next section we show that this male default is encoded in the gate intercepts as well.

\paragraph{Stereotypical referents}
The intermediate conclusion that the language model performs successful anaphora resolution on our experiment also provides us the opportunity to probe the gender biases of the model.
To do so, we repeat the pronoun preference test on an adapted version of the WinoBias corpus \cite{zhao2018gender}, in which all referents are only stereotypically considered to be male or female (e.g., \textit{doctor} and \textit{nurse}).

The results, plotted in Figure~\ref{fig:attention_stereo}, show that the model is very susceptible to stereotypically male referents; these decomposed scores contain an even stronger male preference than for the unambiguous corpus.
The stereotypically female referents, on the other hand, do not lead to a female preference, indicating that their contribution is not considered strong enough evidence by the model to prefer a female pronoun.
All the intermediate tokens exhibit a slight male preference, a pattern that is comparable to the singular bias of the NA task.
From these results we conclude that the model considers a stereotypically male job occupation to be male (``\textit{doctors} are male''), whereas this does not hold for stereotypically female jobs.

\subsection{Pruning information}
Following our subject-verb agreement setup, we compare the predictions of our language model when it focuses only on the subject or object of the sentence.
In Table \ref{tab:gender_accs}, we show the percentage of cases in which \textit{he} is assigned a higher decomposed score than \textit{she}, for both unambiguously gendered referents and stereotypically gendered referents.

\paragraph{\textsc{full}} In the first column of Table~\ref{tab:preds_unamb}, we see that if the sentence contains referents of the same gender (MM \& FF), the full model prediction almost always prefers to use a pronoun with that same gender.
When both a male and female referent are present, the model has a slight preference for generating a pronoun that matches with the \textit{subject} of the sentence (which, interestingly, is the referent that is the furthest away from the pronoun).
In the stereotypical case (Table~\ref{tab:preds_stereo}), the difference between male and female sentences for the \textsc{full} scores almost disappears, showing a predominant male pronoun preference.
This shows that the model by default prefers a masculine pronoun, and only when it is provided sufficient evidence of a female entity it will consider predicting \textit{she} (similar to number agreement).

\paragraph{Pruning} When considering the decompositions with relation to the subject or object we see that the decomposed score of a male entity in all conditions always prefers a male pronoun.
For female entities this effect is slightly obscured by the male bias of the decoder intercept: The accuracies without adding this intercept highlight that female contributions lead to a strong female preference.
For the stereotypical corpus this female preference is far less apparent, which is in line with the results of Section~\ref{sec:gender-exp-1}.
When solely considering the intercept contributions it becomes clear once more that a strong male bias is encoded in them, an effect that is further amplified by the decoder intercept.

\paragraph{Corpus frequency}
For NA the default class turned out to be less frequent in the training corpus.
For our gender setup it turns out the male default is in fact the majority class, with \textit{he} being nearly 4 times more frequent than \textit{she}.
We conclude that the default class is not directly correlated to training frequency and likely depends on the phenomenon at hand, although an investigation incorporating a wider range of models would be needed to establish this.

\section{Conclusion}
We propose a generalised version of Contextual Decomposition \cite{Murdoch2018BeyondLSTMs} -- \ourCD\ -- that allows to study specifically selected interactions of components in an LSTM language model.
This enables \ourCD to extract the contributions of a model's intercepts, or to investigate the interactions of a phrase with other phrases and intercepts. 

We analyse two linguistic phenomena in a pre-trained language model: subject-verb agreement, in which \textit{number} plays a role, and anaphora resolution for which \textit{gender} is important.
Anaphora resolution in the context of language modelling had not been investigated thoroughly before, and our setup enables this at an unprecedented level.

We trace what information the language model uses to make predictions that require gender and number information and find that, in both cases, the model applies a form of \textit{default reasoning}, by falling back on a default class (male, singular) and predicting a female or plural token only when it is provided enough explicit evidence.
As such, the decision to predict masculine and singular words can not be traced back evidently to specific information in the network inputs, but is encoded by default in the model's weights.

Our setup and results demonstrate the power of \ourCD, which can be applied on top of any model without additional training. 
Our results bear relevance for work on detecting and removing model biases, and may clarify some of the issues that were raised by \citet{gonen2019lipstick}, who argue that current bias removal methods only operate on a superficial level.
\ourCD could also be used to aid a model in guiding it towards the right flow of information, which could be applied to a wide range of applications such as the interventions of \citet{Giulianelli2018UnderInformation}. 
In the future, we plan on extending \ourCD to other types of language models, such as the currently popular attention-based models.
Furthermore, we wish to expand the capacities of \ourCD by improving the gate factorisation with a better Shapley value approximator, such as those proposed by \citet{Lundberg2017APredictions} or \citet{ancona2019explaining}.
The axiomatic approach of \citet{Montavon2019} could provide further insight into how \ourCD relates to other explanation methods, and we are confident that combining the strengths of \ourCD with that of other frameworks will ultimately lead to a more \textit{robust} and \textit{faithful} insight into deep neural networks.

\section*{Acknowledgements}

We thank our anonymous reviewers for their useful suggestions and comments. 
DH and WZ are funded by the Netherlands Organization for Scientific Research (NWO), through a Gravitation Grant 024.001.006 to the Language in Interaction Consortium.

\bibliography{references}

\begin{thebibliography}{23}
\expandafter\ifx\csname natexlab\endcsname\relax\def\natexlab#1{#1}\fi

\bibitem[{Ancona et~al.(2019)Ancona, {\"O}ztireli, and
  Gross}]{ancona2019explaining}
Marco Ancona, Cengiz {\"O}ztireli, and Markus Gross. 2019.
\newblock Explaining deep neural networks with a polynomial time algorithm for
  shapley values approximation.
\newblock In \emph{36th International Conference on Machine Learning (ICML
  2019)}.

\bibitem[{Arras et~al.(2017)Arras, Montavon, M{\"u}ller, and
  Samek}]{arras2017explaining}
Leila Arras, Gr{\'e}goire Montavon, Klaus-Robert M{\"u}ller, and Wojciech
  Samek. 2017.
\newblock Explaining recurrent neural network predictions in sentiment
  analysis.
\newblock \emph{EMNLP 2017}, page 159.

\bibitem[{Bach et~al.(2015)Bach, Binder, Montavon, Klauschen, M{\"u}ller, and
  Samek}]{bach2015pixel}
Sebastian Bach, Alexander Binder, Gr{\'e}goire Montavon, Frederick Klauschen,
  Klaus-Robert M{\"u}ller, and Wojciech Samek. 2015.
\newblock On pixel-wise explanations for non-linear classifier decisions by
  layer-wise relevance propagation.
\newblock \emph{PloS one}, 10(7):e0130140.

\bibitem[{Belinkov and Glass(2019)}]{belinkov2019analysis}
Yonatan Belinkov and James Glass. 2019.
\newblock Analysis methods in neural language processing: A survey.
\newblock \emph{Transactions of the Association for Computational Linguistics},
  7:49--72.

\bibitem[{Giulianelli et~al.(2018)Giulianelli, Harding, Mohnert, Hupkes, and
  Zuidema}]{Giulianelli2018UnderInformation}
Mario Giulianelli, Jack Harding, Florian Mohnert, Dieuwke Hupkes, and Willem
  Zuidema. 2018.
\newblock \href {https://doi.org/arXiv:1808.08079v1} {{Under the Hood: Using
  Diagnostic Classifiers to Investigate and Improve how Language Models Track
  Agreement Information}}.
\newblock In \emph{Proceedings of the 2018 EMNLP Workshop BlackboxNLP:
  Analyzing and Interpreting Neural Networks for NLP}, pages 240--248.

\bibitem[{Gonen and Goldberg(2019)}]{gonen2019lipstick}
Hila Gonen and Yoav Goldberg. 2019.
\newblock \href {https://www.aclweb.org/anthology/N19-1061/} {Lipstick on a
  pig: Debiasing methods cover up systematic gender biases in word embeddings
  but do not remove them}.
\newblock In \emph{Proceedings of the 2019 Conference of the North American
  Chapter of the Association for Computational Linguistics: Human Language
  Technologies, {NAACL-HLT} 2019, Minneapolis, MN, USA, June 2-7, 2019, Volume
  1 (Long and Short Papers)}, pages 609--614.

\bibitem[{Gulordava et~al.(2018)Gulordava, Bojanowski, Grave, Linzen, and
  Baroni}]{Gulordava2018ColorlessHierarchically}
Kristina Gulordava, Piotr Bojanowski, Edouard Grave, Tal Linzen, and Marco
  Baroni. 2018.
\newblock \href {http://arxiv.org/abs/1803.11138} {{Colorless green recurrent
  networks dream hierarchically}}.
\newblock In \emph{Proceedings of the 2018 Conference of the North American
  Chapter of the Association for Computational Linguistics: Human Language
  Technologies}, volume~1, pages 1195--1205.

\bibitem[{Hochreiter and Schmidhuber(1997)}]{Hochreiter1997LongMemory}
Sepp Hochreiter and Jürgen Schmidhuber. 1997.
\newblock \href {https://doi.org/10.1162/neco.1997.9.8.1735} {{Long Short-Term
  Memory}}.
\newblock \emph{Neural Computation}, 9(8):1735--1780.

\bibitem[{J{\'{o}}zefowicz et~al.(2016)J{\'{o}}zefowicz, Vinyals, Schuster,
  Shazeer, and Wu}]{googlelm}
Rafal J{\'{o}}zefowicz, Oriol Vinyals, Mike Schuster, Noam Shazeer, and Yonghui
  Wu. 2016.
\newblock \href {http://arxiv.org/abs/1602.02410} {Exploring the limits of
  language modeling}.
\newblock \emph{CoRR}, abs/1602.02410.

\bibitem[{Jumelet and Hupkes(2018)}]{Jumelet2018DoItems}
Jaap Jumelet and Dieuwke Hupkes. 2018.
\newblock {Do Language Models Understand Anything? On the Ability of LSTMs to
  Understand Negative Polarity Items}.
\newblock \emph{Proceedings of the 2018 EMNLP Workshop BlackboxNLP: Analyzing
  and Interpreting Neural Networks for NLP}, pages 222--231.

\bibitem[{Jumelet and Hupkes(2019)}]{diagnnose}
Jaap Jumelet and Dieuwke Hupkes. 2019.
\newblock \href {https://doi.org/10.5281/zenodo.3445477} {diagnnose: A neural
  net analysis library}.

\bibitem[{Lakretz et~al.(2019)Lakretz, Kruszewski, Desbordes, Hupkes, Dehaene,
  and Baroni}]{Lakretz2019TheModels}
Yair Lakretz, German Kruszewski, Theo Desbordes, Dieuwke Hupkes, Stanislas
  Dehaene, and Marco Baroni. 2019.
\newblock {The emergence of number and syntax units in LSTM language models}.
\newblock In \emph{Proceedings of NAACL-HTL 2019}, pages 11--20. Association
  for Computational Linguistics.

\bibitem[{Linzen et~al.(2016)Linzen, Dupoux, and
  Goldberg}]{Linzen2016AssessingDependencies}
Tal Linzen, Emmanuel Dupoux, and Yoav Goldberg. 2016.
\newblock \href {http://arxiv.org/abs/1611.01368} {{Assessing the Ability of
  LSTMs to Learn Syntax-Sensitive Dependencies}}.
\newblock \emph{TACL}, 4:521--535.

\bibitem[{Lundberg and Lee(2017)}]{Lundberg2017APredictions}
Scott~M. Lundberg and Su{-}In Lee. 2017.
\newblock \href
  {http://papers.nips.cc/paper/7062-a-unified-approach-to-interpreting-model-predictions}
  {A unified approach to interpreting model predictions}.
\newblock In \emph{Advances in Neural Information Processing Systems 30: Annual
  Conference on Neural Information Processing Systems 2017, 4-9 December 2017,
  Long Beach, CA, {USA}}, pages 4765--4774.

\bibitem[{Marvin and Linzen(2018)}]{Marvin2018TargetedModels}
Rebecca Marvin and Tal Linzen. 2018.
\newblock \href {http://arxiv.org/abs/1808.09031} {{Targeted Syntactic
  Evaluation of Language Models}}.
\newblock In \emph{EMNLP}, pages 1192--1202. Association for Computational
  Linguistics.

\bibitem[{Montavon(2019)}]{Montavon2019}
Gr{\'e}goire Montavon. 2019.
\newblock \href {https://doi.org/10.1007/978-3-030-28954-6_13}
  {\emph{Gradient-Based Vs. Propagation-Based Explanations: An Axiomatic
  Comparison}}, pages 253--265. Springer International Publishing, Cham.

\bibitem[{Murdoch et~al.(2018)Murdoch, Liu, and Yu}]{Murdoch2018BeyondLSTMs}
W.~James Murdoch, Peter~J. Liu, and Bin Yu. 2018.
\newblock \href {https://openreview.net/forum?id=rkRwGg-0Z} {Beyond word
  importance: Contextual decomposition to extract interactions from lstms}.
\newblock In \emph{6th International Conference on Learning Representations,
  {ICLR} 2018, Vancouver, BC, Canada, April 30 - May 3, 2018, Conference Track
  Proceedings}.

\bibitem[{Olah(2015)}]{colah2015}
Christopher Olah. 2015.
\newblock Understanding lstm networks.
\newblock \url{https://colah.github.io/posts/2015-08-Understanding-LSTMs/}.

\bibitem[{Poerner et~al.(2018)Poerner, Roth, and
  Sch\"utze}]{poernerRothSchutze18acl}
Nina Poerner, Benjamin Roth, and Hinrich Sch\"utze. 2018.
\newblock \href
  {http://nbn-resolving.de/urn/resolver.pl?urn=nbn:de:bvb:19-epub-61866-4}
  {Evaluating neural network explanation methods using hybrid documents and
  morphosyntactic agreement}.
\newblock In \emph{Proceedings of the 56th Annual Meeting of the Association
  for Computational Linguistics (Volume 1: Long Papers)}, pages 340--350,
  Stroudsburg, PA. Association for Computational Linguistics (ACL).

\bibitem[{Shapley(1953)}]{shapley1953value}
Lloyd~S. Shapley. 1953.
\newblock A value for n-person games.
\newblock \emph{Contributions to the Theory of Games}, (28):307--317.

\bibitem[{Singh et~al.(2019)Singh, Murdoch, and
  Yu}]{Singh2019HierarchicalPredictions}
Chandan Singh, W.~James Murdoch, and Bin Yu. 2019.
\newblock \href {http://arxiv.org/abs/1806.05337} {{Hierarchical
  interpretations for neural network predictions}}.
\newblock In \emph{ICLR}.

\bibitem[{Wilcox et~al.(2018)Wilcox, Levy, Morita, and
  Futrell}]{Wilcox2018WhatDependencies}
Ethan Wilcox, Roger Levy, Takashi Morita, and Richard Futrell. 2018.
\newblock \href {http://arxiv.org/abs/1809.00042} {{What do RNN Language Models
  Learn about Filler-Gap Dependencies?}}
\newblock \emph{Proceedings of the 2018 EMNLP Workshop BlackboxNLP: Analyzing
  and Interpreting Neural Networks for NLP}, pages 211--221.

\bibitem[{Zhao et~al.(2018)Zhao, Wang, Yatskar, Ordonez, and
  Chang}]{zhao2018gender}
Jieyu Zhao, Tianlu Wang, Mark Yatskar, Vicente Ordonez, and Kai-Wei Chang.
  2018.
\newblock Gender bias in coreference resolution: Evaluation and debiasing
  methods.
\newblock In \emph{Proceedings of the 2018 Conference of the North American
  Chapter of the Association for Computational Linguistics: Human Language
  Technologies, Volume 2 (Short Papers)}, pages 15--20.

\end{thebibliography}
\bibliographystyle{acl_natbib}

\vfill\break
\newpage
\appendix
\begin{figure*}[h]
\section{Fixed Shapley results -- number agreement}
    \centering
    \begin{subfigure}{0.49\linewidth}
        \centering
        \includegraphics[width=\textwidth, trim=0 0 0mm 0, clip]{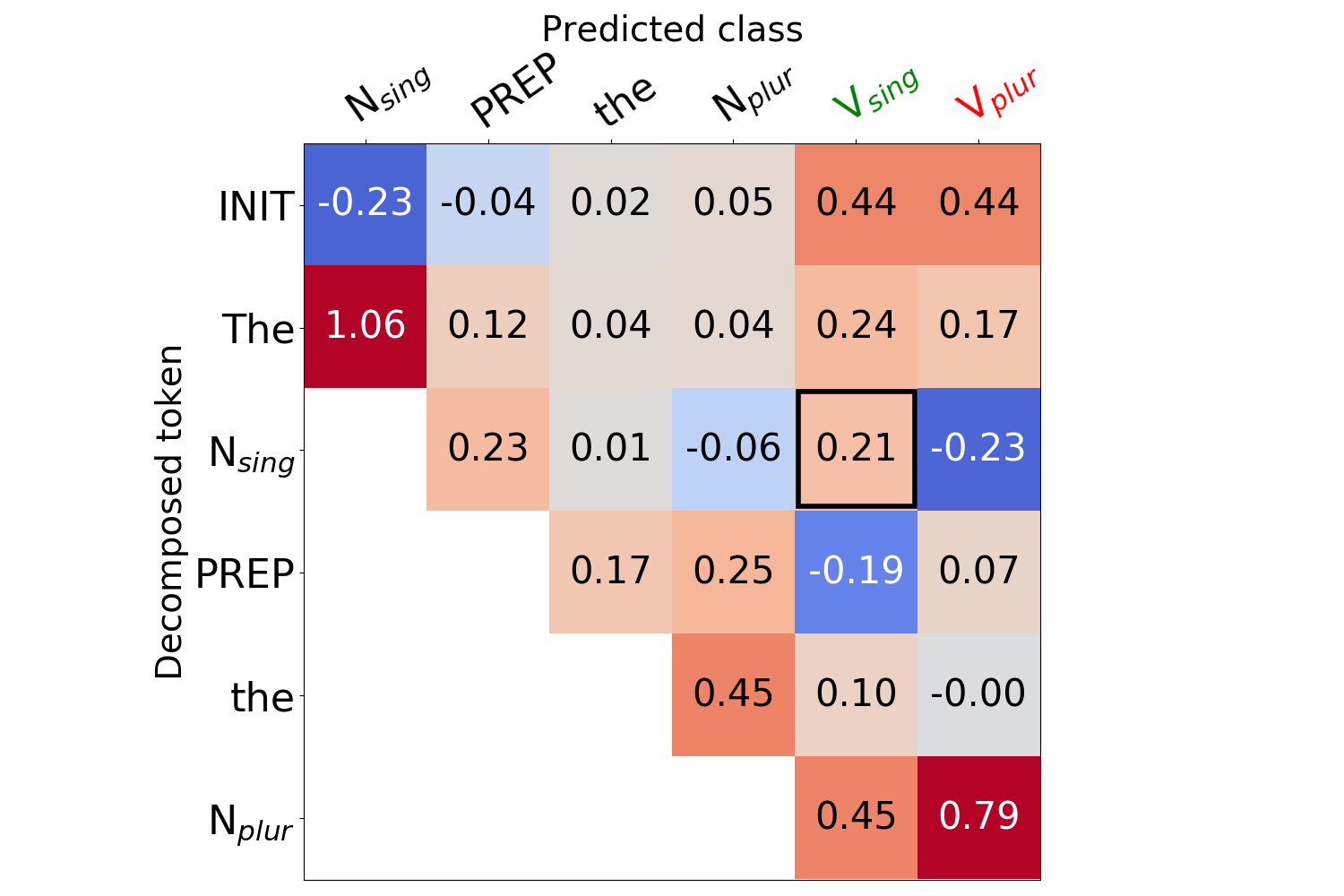}
        \caption{NounPP SP -- \textbf{fixed} Shapley}
    \end{subfigure}
    \hfill
    \begin{subfigure}{0.49\linewidth}
        \centering
        \includegraphics[width=\textwidth, trim=0 0 0mm 0, clip]{figures/nounSP_new.png}
        \caption{NounPP SP-- \textbf{full} Shapley}
    \end{subfigure}\\[5pt]
    \begin{subfigure}{0.49\linewidth}
        \centering
        \includegraphics[width=\textwidth, trim=0 0 0mm 0, clip]{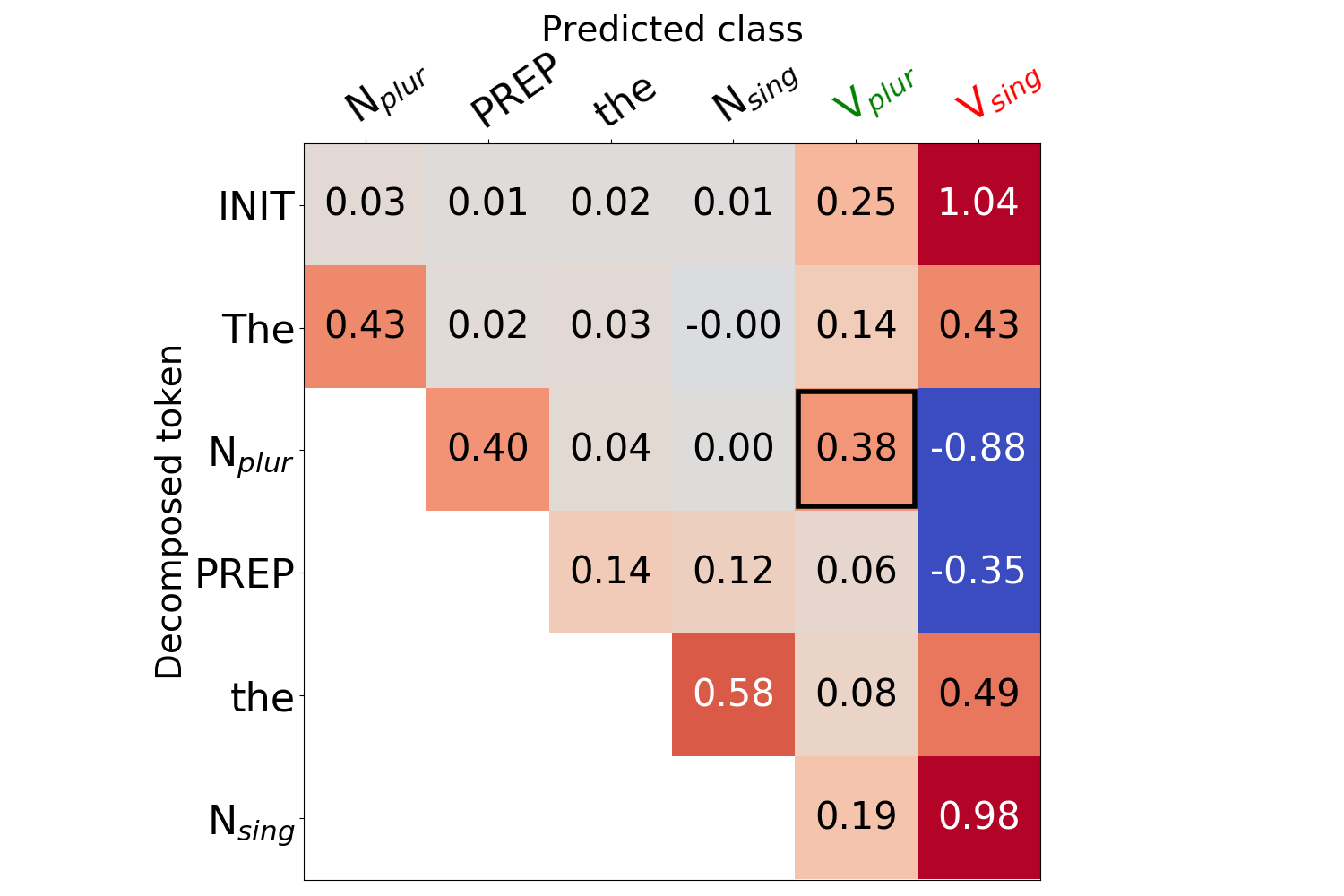}
        \caption{NounPP PS -- \textbf{fixed} Shapley}
    \end{subfigure}
    \hfill
    \begin{subfigure}{0.49\linewidth}
        \centering
        \includegraphics[width=\textwidth, trim=0 0 0mm 0, clip]{figures/nounPS_new.png}
        \caption{NounPP PS -- \textbf{full} Shapley}
    \end{subfigure}
    \caption{Results of Figure~\ref{fig:sv_cd_matrix}, for both Shapley computations. Note how the fixed Shapley results generally lead to lower term contributions, as these are more prominently assigned to the intercept terms instead.}\label{fig:full_sv_cd_matrix}
\end{figure*}

\begin{table*}[h]
\footnotesize
\centering
\setlength{\tabcolsep}{3pt}
\begin{tabular}{|c|c||c|ccc|}
\cline{4-6}\multicolumn{3}{c}{} & \multicolumn{3}{|c|}{\Tstrut \textsc{gcd} -- \textbf{fixed} Shapley} \\
\hline\textbf{NA Task} & \textbf{C} & \textsc{full} & \textsc{in} & \textsc{intercept}$^*$ & \textsc{$\neg$intercept}\Tstrut\\\hline
\textcolor{persiangreen}{Simple}    & \textcolor{persiangreen}{S}  & \textcolor{persiangreen}{100}  & \textcolor{persiangreen}{100}  & \textcolor{persiangreen}{100} & \textcolor{persiangreen}{7.7}  \Tstrut\\
Simple             & P           & 100           & 100           & 7.3          & 65.7  \Bstrut\\\hline\Tstrut
\textcolor{persiangreen}{nounPP}    & \textcolor{persiangreen}{SS} & \textcolor{persiangreen}{99.2} & \textcolor{persiangreen}{91.2} & \textcolor{persiangreen}{100} & \textcolor{persiangreen}{14.8}  \\
\textcolor{persiangreen}{nounPP}    & \textcolor{persiangreen}{SP} & \textcolor{persiangreen}{87.2} & \textcolor{persiangreen}{91.7} & \textcolor{persiangreen}{100} & \textcolor{persiangreen}{14.3}  \\
nounPP             & PS          & 92.0          & 100           & 0            & 82.7  \\
nounPP             & PP          & 99.0          & 99.8           & 0.5          & 81.0  \Bstrut\\\hline\Tstrut
\textcolor{persiangreen}{namePP}    & \textcolor{persiangreen}{SS} & \textcolor{persiangreen}{99.3} & \textcolor{persiangreen}{91.2} & \textcolor{persiangreen}{100} & \textcolor{persiangreen}{12.4}  \\
namePP             & PS          & 68.9          & 99.8          & 0            & 82.0  \Bstrut\\\hline
\end{tabular}\hfill
\begin{tabular}{|c|c||c|ccc|}
\cline{4-6}\multicolumn{3}{c}{} & \multicolumn{3}{|c|}{\Tstrut \textsc{gcd} -- \textbf{full} Shapley} \\
\hline\textbf{Task} & \textbf{C} & \textsc{full} & \textsc{in} & \textsc{intercept}$^*$ & $\neg$\textsc{intercept}\Tstrut\\\hline\Tstrut
\textcolor{persiangreen}{Simple}    & \textcolor{persiangreen}{S}  & \textcolor{persiangreen}{100} & \textcolor{persiangreen}{73.3} \scriptsize{(91.3)} & \textcolor{persiangreen}{97.3} \scriptsize{(100)} & \textcolor{persiangreen}{69.7} \scriptsize{(86.3)} \Tstrut\\
Simple             & P           & 100        & 100  \scriptsize{(100)}          & 32.7 \scriptsize{(7.7)}         & 100 \scriptsize{(100)}  \Bstrut\\\hline\Tstrut
\textcolor{persiangreen}{nounPP}    & \textcolor{persiangreen}{SS} & \textcolor{persiangreen}{99.2} & \textcolor{persiangreen}{93.0} \scriptsize{(99.7)}& \textcolor{persiangreen}{99.8} \scriptsize{(99.8)} & \textcolor{persiangreen}{72.7} \scriptsize{(88.7)} \\
\textcolor{persiangreen}{nounPP}    & \textcolor{persiangreen}{SP} & \textcolor{persiangreen}{87.2} & \textcolor{persiangreen}{90.3} \scriptsize{(99.3)} & \textcolor{persiangreen}{98.8} \scriptsize{(99.8)} & \textcolor{persiangreen}{60.5} \scriptsize{(83.5)} \\
nounPP             & PS          & 92.0          & 100   \scriptsize{(100)}        & 0.0   \scriptsize{(0.0)}         & 100 \scriptsize{(100)} \\
nounPP             & PP          & 99.0          & 100   \scriptsize{(99.3)}        & 7.0 \scriptsize{(0.5)}         & 99.8  \scriptsize{(100)} \Bstrut\\\hline\Tstrut
\textcolor{persiangreen}{namePP}    & \textcolor{persiangreen}{SS} & \textcolor{persiangreen}{99.3} & \textcolor{persiangreen}{97.7} \scriptsize{(91.3)} & \textcolor{persiangreen}{99.4} \scriptsize{(100)} & \textcolor{persiangreen}{76.2} \scriptsize{(90.9)} \\
namePP             & PS          & 68.9          & 98.3 \scriptsize{(98.2)}          & 1.3 \scriptsize{(0.0)}            & 99.9 \scriptsize{(99.9)}  \Bstrut\\\hline
\end{tabular}
\caption{
Results of Table~\ref{tab:sv-accuracies}, for both Shapley computations. 
The main difference here lies in the $\neg$\textsc{intercept} case: for the fixed Shapley this case leads to a much starker decrease.
The pattern, however, remains unaltered: the singular conditions depend much stronger on the intercepts than the plural conditions for both the Shapley computations.
}
\end{table*}

\begin{figure*}[h]
\section{Fixed Shapley results -- pronoun resolution}
\centering
        \captionsetup[subfigure]{justification=centering}
    \begin{subfigure}{0.24\linewidth}
        \centering
        \includegraphics[width=\textwidth, trim=0 0 0mm 0, clip]{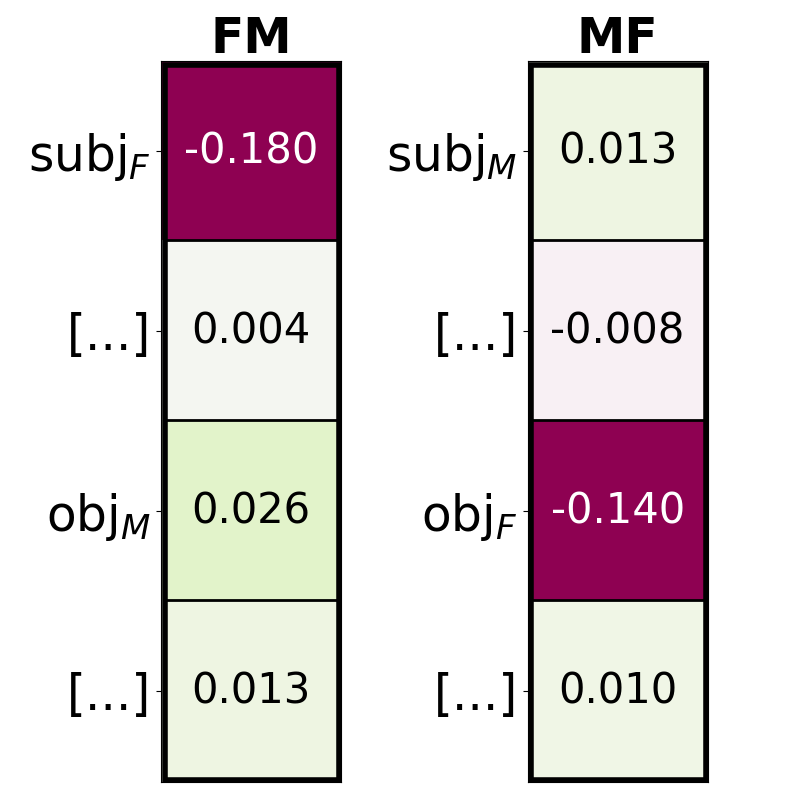}
        \caption{unambiguous\\\textbf{fixed} Shapley}
    \end{subfigure}
    \qquad~~
    \begin{subfigure}{0.24\linewidth}
        \centering
        \includegraphics[width=\textwidth, trim=0 0 0mm 0, clip]{figures/unamb.png}
        \caption{unambiguous\\ \textbf{full} Shapley}
    \end{subfigure}
    \\[10pt]
    \begin{subfigure}{0.24\linewidth}
        \centering
        \includegraphics[width=\textwidth, trim=0 0 0mm 0, clip]{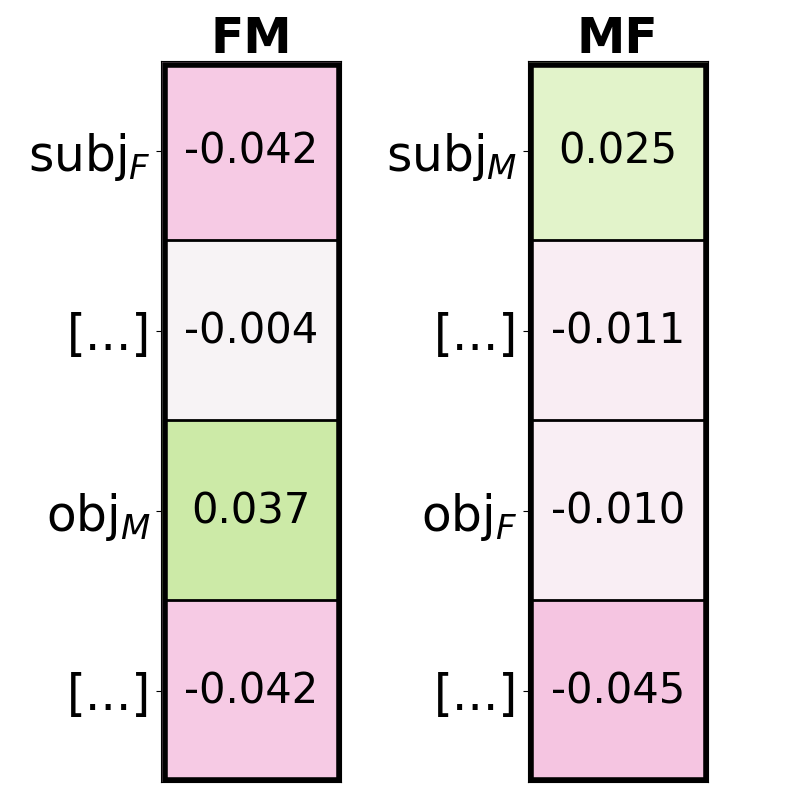}
        \caption{stereotypical\\ \textbf{fixed} Shapley}
    \end{subfigure}
    \qquad~~
    \begin{subfigure}{0.24\linewidth}
        \centering
        \includegraphics[width=\textwidth, trim=0 0 0mm 0, clip]{figures/stereo.png}
        \caption{stereotypical\\ \textbf{full} Shapley}
    \end{subfigure}
\caption{
Results of Figure~\ref{fig:gender_attention}, for both Shapley computations.
The pattern remains the same, although the full Shapley case highlights a stronger default male bias that is encoded in the non-gendered sub-phrases.
}\label{fig:full_gender_attention}
\end{figure*}

\begin{table*}[h]
\footnotesize
\centering
\begin{subtable}[t]{0.48\textwidth}
\centering
    \begin{tabular}{|c||c|ccc|}
    \cline{3-5}\multicolumn{2}{c}{} & \multicolumn{3}{|c|}{\Tstrut \textsc{gcd} -- \textbf{fixed} Shapley} \\
        \hline
        \textbf{C} & \textsc{full} & \textsc{subject} & \textsc{object} & \textsc{intercept} \Tstrut\Bstrut\\
        \hline
        MM  & 100 & 100 & 100 & 100 \Tstrut\\
        MF  & 58.6 & 100 & 31.2 & 100 \\
        FM  & 37.0 & 6.2 & 100 & 100\\
        FF  & 1.2 & 50.0 & 73.6 & 100\\
        \hline
    \end{tabular}
    \caption{\%\textit{he}\textgreater \textit{she}, unambiguous referents}
\end{subtable}
\quad
\begin{subtable}[t]{0.48\textwidth}
\centering
    \begin{tabular}{|c||c|ccc|}
    \cline{3-5}\multicolumn{2}{c}{} & \multicolumn{3}{|c|}{\Tstrut \textsc{gcd} -- \textbf{full} Shapley} \\
        \hline
        \textbf{C} & \textsc{full} & \textsc{subject} & \textsc{object} & \textsc{intercept}$^*$ \Tstrut\Bstrut\\
        \hline
        MM  & 100 & 100 \scriptsize{(93.2)} & 100 \scriptsize{(97.8)} & 100 \scriptsize{(93.2)} \Tstrut\\
        MF  & 58.6 & 100 \scriptsize{(86.4)} & 47.2 \scriptsize{(0.8)} & 100 \scriptsize{(96.0)} \\
        FM  & 37.0 & 29.2 \scriptsize{(0.6)} & 100 \scriptsize{(97.2)} & 100 \scriptsize{(98.0)}\\
        FF  & 1.2 & 77.2 \scriptsize{(0.8)} & 88.8 \scriptsize{(1.2)} & 100 \scriptsize{(92.2)}\\
        \hline
    \end{tabular}
    \caption{\%\textit{he}\textgreater \textit{she}, unambiguous referents}
\end{subtable}
\\[10pt]

\begin{subtable}[t]{0.48\textwidth}
\centering
    \begin{tabular}{|c||c|ccc|}
    \cline{3-5}\multicolumn{2}{c}{} & \multicolumn{3}{|c|}{\Tstrut \textsc{gcd} -- \textbf{fixed} Shapley} \\
        \hline
        \textbf{C} & \textsc{full} & \textsc{subject} & \textsc{object} & \textsc{intercept} \Tstrut\Bstrut\\
        \hline
        MM  & 100 & 100 & 100 & 100 \Tstrut\\
        MF  & 94.6 & 100 & 89.4 & 100 \\
        FM  & 88.8 & 81.6 & 100 & 100\\
        FF  & 84.6 & 83.0 & 92.2 & 100\\
        \hline
    \end{tabular}
    \caption{\%\textit{he}\textgreater \textit{she}, stereotypical referents}
\end{subtable}
\quad
\begin{subtable}[t]{0.48\textwidth}
\centering
    \begin{tabular}{|c||c|ccc|}
    \cline{3-5}\multicolumn{2}{c}{} & \multicolumn{3}{|c|}{\Tstrut \textsc{gcd} -- \textbf{full} Shapley} \\
        \hline
        \textbf{C} & \textsc{full} & \textsc{subject} & \textsc{object} & \textsc{intercept}$^*$ \Tstrut\Bstrut\\
        \hline
        MM  & 100 & 100 \scriptsize{(100)} & 100 \scriptsize{(100)} & 100 \scriptsize{(88.0)}\Tstrut\\
        MF  & 94.6 & 100 \scriptsize{(99.6)} & 95.4 \scriptsize{(84.0)} & 100 \scriptsize{(84.8)}\\
        FM  & 88.8 & 90.6 \scriptsize{(77.4)} & 100 \scriptsize{(100)} & 100 \scriptsize{(91.0)}\\
        FF  & 84.6 & 92.8 \scriptsize{(75.6)} & 97.4 \scriptsize{(84.0)} & 100 \scriptsize{(89.2)}\\
        \hline
    \end{tabular}
    \caption{\%\textit{he}\textgreater \textit{she}, stereotypical referents}
\end{subtable}

\caption{
Results of Table~\ref{tab:gender_accs}, for both Shapley computations. Similar to Figure~\ref{fig:full_gender_attention}, it can be seen that the pattern remains the same, with the full Shapley computation again highlighting a slightly stronger male bias.
}\vspace{-4mm}
\end{table*}

\end{document}